\documentclass[sigconf, nonacm]{acmart}
\AtBeginDocument{%
  }

\usepackage{array}
\usepackage{graphicx}
\usepackage{latexsym}
\usepackage{amsthm}
\usepackage{lscape}
\usepackage{booktabs}
\usepackage{enumitem}
\usepackage{graphicx}
\usepackage{color}
\usepackage{soul}
\usepackage{algorithmicx}
\usepackage[T1]{fontenc}
\usepackage{multirow}
\usepackage{algorithm, algorithmicx, algpseudocode}
\usepackage{amsmath} 
\usepackage{bm} 
\usepackage[misc]{ifsym}
\usepackage{natbib}
\usepackage{balance}
\usepackage{enumitem}
\usepackage{caption}  
\usepackage{hyperref}
\usepackage{float}

\begin{document}

\title{ITL-LIME: Instance-Based Transfer Learning for Enhancing Local Explanations in Low-Resource Data Settings}

\thanks{This paper has been accepted at the 34th ACM International Conference on Information and Knowledge Management (CIKM 2025).}

\author{Rehan Raza}
\orcid{0000-0002-2938-8470}
\affiliation{%
  \department{School of Information Technology}
  \institution{Murdoch University}
  \city{Perth}
  \state{Western Australia}
  \country{Australia}
}
\email{35021174@student.murdoch.edu.au}

\author{Guanjin Wang}
\orcid{0000-0002-5258-0532}
\affiliation{%
  \department{School of Information Technology}
  \institution{Murdoch University}
  \city{Perth}
  \state{Western Australia}
  \country{Australia}
  }
\email{Guanjin.Wang@murdoch.edu.au}
\authornote{Corresponding author.}

\author{Kok Wai Wong}
\orcid{0000-0001-8767-1031}
\affiliation{%
  \department{School of Information Technology}
  \institution{Murdoch University}
  \city{Perth}
  \state{Western Australia}
  \country{Australia}
}
\email{K.Wong@murdoch.edu.au} 

\author{Hamid Laga}
\orcid{0000-0002-4758-7510}
\affiliation{%
  \department{School of Information Technology}
  \institution{Murdoch University}
  \city{Perth}
  \state{Western Australia}
  \country{Australia}
  }
\email{h.laga@murdoch.edu.au} 

\author{Marco Fisichella}
\orcid{0000-0002-6894-1101}
\affiliation{%
  \department{L3S Research Center}
  \institution{Leibniz University}
  \city{Hannover}
  \country{Germany}}
\email{mfisichella@l3s.de}

\renewcommand{\shortauthors}{Rehan Raza, Guanjin Wang, Kok Wai Wong, Hamid Laga, Marco Fisichella}

\begin{abstract}
Explainable Artificial Intelligence (XAI) methods, such as Local Interpretable Model-Agnostic Explanations (LIME), have advanced the interpretability of black-box machine learning models by approximating their behavior locally using interpretable surrogate models. 
However, LIME's inherent randomness in perturbation and sampling can lead to locality and instability issues, especially in scenarios with limited training data. In such cases, data scarcity can result in the generation of unrealistic variations and samples that deviate from the true data manifold. Consequently, the surrogate model may fail to accurately approximate the complex decision boundary of the original model. 
To address these challenges, we propose a novel Instance-based Transfer Learning LIME framework (ITL-LIME) that enhances explanation fidelity and stability in data-constrained environments. 
ITL-LIME introduces instance transfer learning into the LIME framework by leveraging relevant real instances from a related source domain to aid the explanation process in the target domain. 
Specifically, we employ clustering to partition the source domain into clusters with representative prototypes. Instead of generating random perturbations, our method retrieves pertinent real source instances from the source cluster whose prototype is most similar to the target instance. These are then combined with the target instance's neighboring real instances. 
To define a compact locality, we further construct a contrastive learning-based encoder as a weighting mechanism to assign weights to the instances from the combined set based on their proximity to the target instance. Finally, these weighted source and target instances are used to train the surrogate model for explanation purposes.
Experimental evaluation with real-world datasets demonstrates that ITL-LIME greatly improves the stability and fidelity of LIME explanations in scenarios with limited data. Our code is available at: \url{https://github.com/rehanrazaa/ITL-LIME}.
\end{abstract}




\maketitle

\section{Introduction}
AI, including ML and DL, is increasingly deployed in high-stakes domains, such as healthcare, finance, and criminal justice  \cite{ribeiro2018anchors}. It is crucial for stakeholders to comprehend the AI decision-making process to improve transparency, accountability, and trust \cite{slack2021reliable}. However, many models operate as "black-boxes", offering limited insight into their internal reasoning \cite{2022breaking}. 
Explainable AI (XAI) has emerged as an important field that aims to demystify these opaque models by providing interpretable explanations of their black-box predictions \cite{10101766}. Local Interpretable Model-Agnostic Explanations (LIME) \cite{ribeiro2016should} is a widely used XAI method that provides local, model-agnostic explanations for individual predictions by constructing an interpretable surrogate model that approximates the behavior of complex models in the vicinity of a specific instance \cite{ribeiro2016should}. 

While LIME has been widely adopted for model explanation, it faces challenges related to locality and instability \cite{sangroya2020guided,knab2025lime}. LIME relies on random perturbations to generate synthetic samples. 
However, when the synthetic samples do not adequately represent the local decision boundary of the black-box model, the resulting explanations suffer from locality issues \cite{laugel2018defining}. In addition, LIME explanations can be unstable due to the inherent randomness in the perturbation and sampling procedures \cite{tan2023glime}. Fidelity is another closely related concern  \cite{sokol2025limetree,yasui2023improving} that arises when the sampling and weighting process fails to generate a meaningful local neighborhood, or when feature representations are inadequate, resulting in local surrogate models that do not faithfully reflect the behavior of the original model \cite{zhou2021s, knab2025lime}.
Furthermore, traditional LIME perturbs the feature of the instance using univariate distributions, which overlooks feature correlations \cite{zhang2019should,bora2024slice}, potentially leading to unrealistic variations \cite{10.1007/978-3-030-28730-6_4,saini2022select}. 
These challenges are further exacerbated in data-scarce settings, e.g., healthcare, where privacy regulations and ethical concerns often limit the availability of rich datasets \cite{10101766}. In such contexts, perturbations are even more likely to generate unrealistic variations and produce samples falling outside the true data manifold. Consequently, the surrogate model trained on these samples may struggle to accurately approximate the local behavior of the black-box model, further degrading the fidelity and stability of the explanations on small datasets.
These limitations are particularly problematic in critical applications where high reliability and trustworthiness of model explanations are important \cite{garreau2020looking,rahnama2024can}. 

To address the limitations of LIME in low-resource data settings, we propose ITL-LIME, a novel instance-based transfer learning framework that enhances LIME's stability and local fidelity by leveraging real instances from a related source domain. 
Unlike traditional LIME, which relies on synthetic perturbations that may produce unrealistic samples, especially with limited data, ITL-LIME constructs explanations using real neighboring instances from both the target and source domains. 
Specifically, our method first applies $K$-medoids clustering to partition the source domain into representative prototype clusters. For a given target instance, ITL-LIME identifies the most similar source prototype and retrieves its member instances, then combines them with real neighboring target instances identified around the instance being explained. 
To establish a compact and meaningful neighborhood around the target instance, we train a contrastive learning-based encoder on the combined set of source and target instances that assigns proximity-based weights by computing distances between the target instance and each instance in the combined set. 
Finally, the weighted source and target instances are used to fit a local surrogate model. 

The main contribution of this study can be summarized as follows:
\begin{itemize} [noitemsep, topsep=0pt]
    \item We introduce ITL-LIME, a novel framework that introduces instance-based transfer learning into LIME to enhance explanation fidelity and stability, particularly in low-resource data settings. 
    \item ITL-LIME constructs a combined set of real instances from both source and target domains that are proximal to the target instance being explained. A contrastive learning-based encoder is then trained on this set to assign proximity-based weights computed in the latent embedding space to refine the locality definition for the surrogate model.   
    \item Extensive experiments on real-world datasets demonstrate that ITL-LIME significantly improves the stability and fidelity of LIME explanations in data-scarce environments, as measured by various quantitative metrics.
\end{itemize}

The remaining sections of this paper are organized as follows: Section \ref{RW} reviews relevant work. Section \ref{PM} details the proposed method. Section \ref{EXP} discusses the experiments and results. Finally, Section \ref{CF} concludes the paper. 

\section{Related work} \label{RW}
\textbf{LIME-based methods.}
LIME \cite{ribeiro2016should} is one of the most widely used post-hoc methods for explaining individual predictions of built machine learning models. Despite its popularity, LIME has notable limitations, including issues with locality, fidelity, and stability, which present significant challenges for reliable interpretation.
To address these issues, various LIME variants have been proposed, focusing on approximation-based \cite{zhou2021s}, selection-based \cite{sangroya2020guided, rabold2018explaining}, distribution-based \cite{meng2024segal, shankaranarayana2019alime}, and neighborhood-based methods \cite{botari2020melime, zafar2019dlime, make3030027, knab2025lime}.
For example, Zhou et al. \cite{zhou2021s} introduced the Stabilize LIME (S-LIME), which uses a hypothesis-testing framework based on the central limit theorem to determine the number of perturbed samples required for stable explanations. Similarly, BayLIME \cite{zhao2021baylime} uses prior knowledge based on Bayesian reasoning to enhance the fidelity and stability.
ALIME \cite{shankaranarayana2019alime}, pre-generate a large set of samples from a Gaussian distribution and utilize a denoising autoencoder as a weighting function. The autoencoder, trained on the training dataset, embeds generated data around the instance of interest and computes distances in the embedded space, leading to a more consistent and locally accurate interpretable model. However, such method's performance can be limited by the original dataset sample size. In low-sample scenarios, the autoencoder may fail to learn generalizable patterns, resulting in unstable latent features and unreliable weightings.
Zafar et al. \cite{zafar2019dlime, make3030027} introduce deterministic LIME, which uses hierarchical clustering to group training data and KNN to identify the cluster closest to the instance being explained. Then it uses the selected cluster's instances to build a simple local model for explanations with improved fidelity and stability. However, such a deterministic LIME model's performance is highly sensitive to dataset size. A small sample may degrade the quality of clustering and consequently affect the explanation's stability and locality in low-data-source settings.
To our knowledge, little work has addressed improving LIME's explainability for black-box models under such constraints. This work introduces the ITL-LIME framework, designed to enhance the stability and fidelity of LIME's local explanations in limited data scenarios. 

\textbf{Instance-based transfer learning (ITL).}
It is a type of transfer learning that focuses on improving model performance in a target domain, usually with limited data, by leveraging specific data instances from a related source domain \cite{zhuang2020comprehensive}. Instead of using the full source dataset or transferring model parameters or features, ITL carefully selects or reweights source domain instances that are most relevant to the target task. These selected source instances help the target model better adapt to the target data distribution. For example, 
Asgarian et al. \cite{asgarian2018hybrid} propose a hybrid ITL framework, which assigns soft weights to source instances by combining task relevance and domain similarity. Domain similarity was estimated using a discriminative classifier trained to differentiate source and target instances, while task relevance was assessed by measuring the model's uncertainty about individual source samples. This hybrid weighting mechanism encourages robust transfer, particularly in limited resource settings. 
Wang et al. \cite{wang2019instance} introduced the ITL method that used a pre-trained model from the source domain to estimate the influence of training samples in the target domain and then optimize the training data by eliminating the samples that negatively influence the performance of the pre-trained model, resulting in improved generalization. 
Most LIME-related works that incorporate transfer learning focus primarily on enhancing the performance of black-box models, with explainability applied post hoc \cite{10100703,arslan2024analysis}. In contrast, this work integrates ITL directly into the LIME framework to explicitly improve local explanation quality on small datasets by leveraging informative instances from a data-rich source domain.

\textbf{Self-supervised learning (SSL).}
It aims to learn meaningful representations from unlabeled data and is typically divided into two main approaches: pretext task-based methods and contrastive learning. Pretext tasks are auxiliary objectives (e.g., predicting masked features or reconstructing input) that generate supervision signals from the data itself. Contrastive learning trains an encoder by generating augmented views of the same input and encouraging similar representations for those views \cite{li2022selective}. While SSL has shown strong performance in computer vision and natural language processing tasks \cite{laine2019high,noroozi2016unsupervised,tian2020contrastive}, its application to tabular data remains relatively limited. Nonetheless, a few recent studies have begun exploring SSL techniques for tabular domains.
For instance, VIME \cite{yoon2020vime} is a semi-supervised, SSL framework for tabular data that uses two pretext tasks—predicting the mask vector and reconstructing the masked features from corrupted input to learn representations without labeled data. However, its effectiveness can degrade when the quality of the generated masks is poor, particularly in the presence of noise or non-uniform feature distributions.
Ucar et al. \cite{ucar2021subtab} introduced SubTab for feature subsetting in tabular data. It divides input features into subsets to learn multi-view representations instead of relying on the corrupted version in an autoencoder setting. 
Bahri et al. \cite{bahriscarf} presented SCARF, a contrastive learning method for tabular data, where the multiple views of the input data are generated by corrupting a random subset of features and replacing them with the values drawn from the feature’s distribution. The method learns the representations by maximizing the similarity between views of the same input. 
Results show that SCARF's effectiveness in downstream classification tasks, even in the presence of limited labeled data or label noise. We used the SCARF \cite{bahriscarf} method to design the weighting for the neighborhood of the instance to explain, including both source and target instances in the proposed ITL-LIME framework. It is selected for its proven effectiveness, surpassing denoising autoencoders, such as those used in VIME \cite{yoon2020vime}, and its stability with respect to hyperparameter settings.
\section{ITL-LIME} \label{PM}
\begin{figure*}[h]
\centering
\includegraphics[width=1 \textwidth]{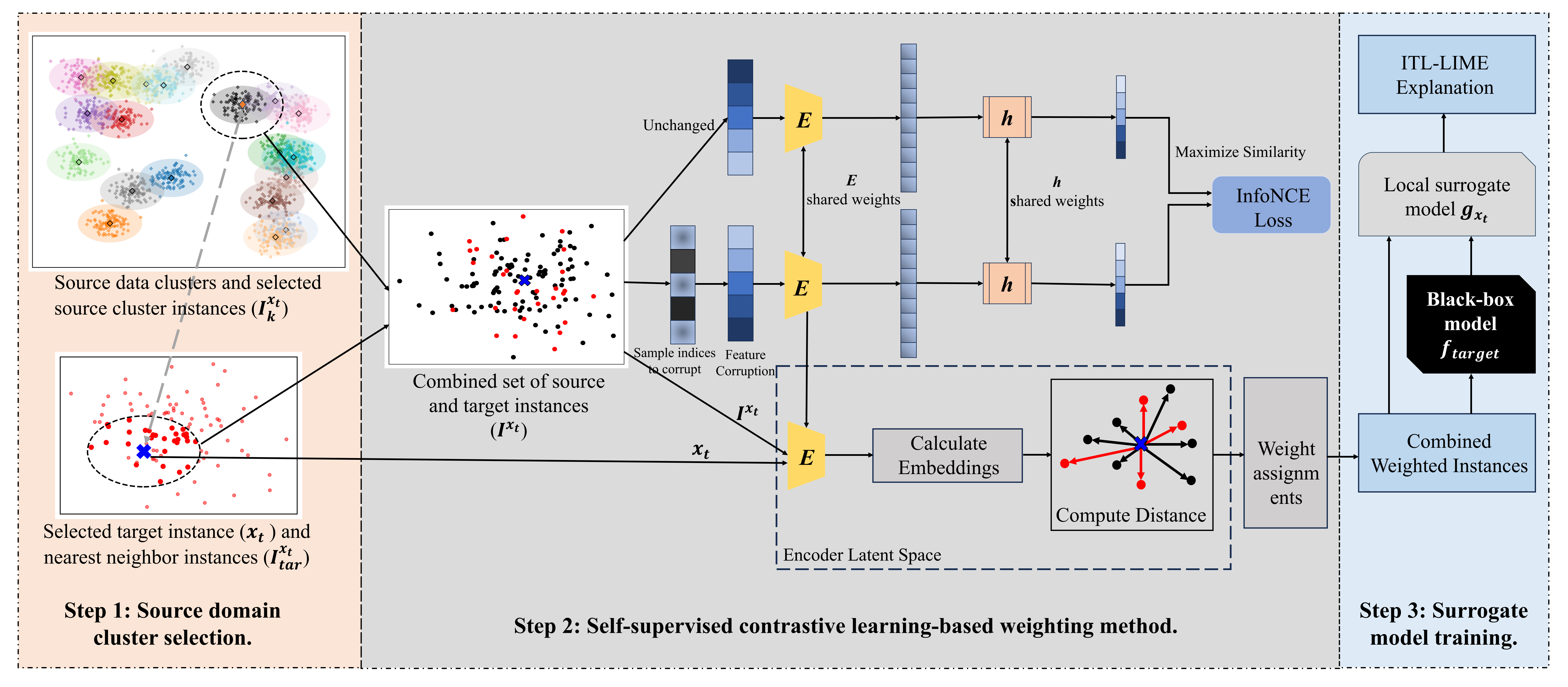}
\caption{Workflow of the proposed ITL-LIME method.}
\label{ssl_flow}
\end{figure*}

We propose a new ITL-LIME framework that enhances LIME’s local fidelity and stability on small datasets (target domain) by leveraging relevant real instances from a related, data-rich source domain containing larger and more diverse samples.
The proposed ITL-LIME framework consists of three main steps. In the first step, it employs \(K\)-medoids clustering to partition the source data into clusters with representative prototypes. Instead of random perturbations, our method selects relevant real source instances from the cluster whose prototype is most similar to the target instance that is being explained. In step $2$, these source instances are combined with the target instance’s neighboring real instances. 
Then, these combined instances are passed to a contrastive learning-based encoder that serves as a weighting mechanism to assign weights to each selected source and target instance based on distance from the target instance being explained. Finally, in step $3$, these weighted combined instances are used to train a simple interpretable model for explanation purposes. 
The following subsections provide details of the training of ITL-LIME. 
The overall workflow of the proposed ITL-LIME framework is illustrated in Figure \ref{ssl_flow}.
\subsection{Source domain cluster selection}
\sloppy
Given a source domain \( \mathcal{D}_S = \{(\bm{x}_i^s, y_i^s)\}_{i=1}^{n_s} \), where each 
\( \bm{x}_i^s = [x_{i1}, x_{i2}, \ldots, x_{id}] \in \mathcal{X}_S = \mathbb{R}^d \), 
\( x_{ij} \in \mathcal{X}_j \) denotes the \( j \)-th individual feature, and 
\( y_i^s \in \mathcal{Y}_S \), our objective in \textbf{Step 1} is to partition the source data into clusters using only unlabeled source inputs. 
We employ the \( K \)-medoids clustering algorithm to obtain a set of representative centroids, where each centroid corresponds to a real data point in the source dataset. 
Unlike methods such as $K$-means, \( K \)-medoids ensures that the clusters are actual data points from the dataset, which is especially important in applications requiring human interpretability, such as healthcare, where real data points can be more directly analyzed and referenced to facilitate explanation procedures. 

\( K \)-medoids seeks to identify the set of centroids \( \mathcal{C} = \{\bm{c}_1, \bm{c}_2, \ldots, \bm{c}_K\} \) and partition the source data into $K$ clusters by minimizing the total dissimilarity within each clusters:
\begin{equation}
    \underset{\{\bm{c}_1,\bm{c}_2,\cdots,\bm{c}_K\}}{\text{min}} \; \sum_{k=1}^K \sum_{\bm{x}^{s} \in \mathcal{I}_k} \text{d}(\bm{x}^{s}, \bm{c}_k),
\label{kmedoid}
\end{equation}
where \( \mathcal{I}_k \) denotes the set of data points assigned to the \(k\)-th cluster \(\bm{c}_k\), and \( \text{d}(\cdot, \cdot) \) is a suitable distance metric. 

\subsection{Instance-based transfer learning for target surrogate model training}
Given a target domain \( \mathcal{D}_T = \{(\bm{x}_i^t, y_i^t)\}_{i=1}^{n_t} \), where each 
\( \bm{x}_i^t = [x_{i1}, x_{i2}, \ldots, x_{id}] \in \mathcal{X}_T = \mathbb{R}^d \), 
\( x_{ij} \in \mathcal{X}_j \) denotes the \( j \)-th individual feature, and 
\( y_i^t \in \mathcal{Y}_T\), and a corresponding trained black-box model $f_{target}$. We assume that the target domain shares the same feature and label spaces as the source domain (i.e, $\mathcal{X}_S = \mathcal{X}_T, \mathcal{Y}_S = \mathcal{Y}_T$), but differs in its input distribution, i.e, \( P_T(\bm{x}) \neq P_S(\bm{x}) \).

In \textbf{Step 2}, to enable instance-based transfer learning, we begin by selecting a target instance of interest $\bm{x}_{t}$, and identifying the most similar source centroid \(\bm{c}_k^{x_t}\) from the source domain by computing the distance, such as Euclidean, cosine, or Gower distance, between $\bm{x}_{t}$ and all source centroids. The centroid \(\bm{c}_k^{x_t}\) is selected as the nearest in distance and label-consistent with  $\bm{x}_{t}$.
Once the appropriate source centroid \(\bm{c}_k^{x_t}\) is selected, the set \(\mathcal{I}_k^{x_t}\) comprising real source instances assigned to \(\bm{c}_k^{x_t}\) is retrieved to support the transfer process.  

Next, to explain the prediction for a given target instance $\bm{x}_{t}$, our method utilizes $k$-nearest neighbors (KNN) approach to identify a set of real, similar neighboring instances to $\bm{x}_{t}$. 
Specifically, KNN computes the distance between $\bm{x}_{t}$ and each target instance $\bm{x}_{i}$ in the dataset, and selects the \(k\) closest neighbors to form \(\mathcal{I}_{tar}^{x_t}\). This set represents the local neighborhood of real target instances surrounding $\bm{x}_{t}$. This neighborhood size or $k$ can vary depending on the overall dataset size and is determined based on $\xi$, defined as the transferred source sample size relative to the target neighborhood size. Based on our experimental trials, we recommend selecting $\xi$ from the values $\{1\!:\!0.5,\ 1\!:\!0.75,\ 1\!:\!1\}$.

We further construct a unified neighborhood set \( \mathcal{I}^{x_t}\) by combining the local target neighborhood \(\mathcal{I}_{tar}^{x_t}\) with the transferred source cluster set \(\mathcal{I}_k^{x_t}\), which is retrieved based on the label-consistent nearest source centroid (\(\bm{c}_k^{x_t}\)), which can be defined as \( \mathcal{I}^{x_t} =\mathcal{I}_{k}^{x_t} \ \cup \ \mathcal{I}_{tar}^{x_t} \).
This unified set enriches the local context for generating explanations by incorporating both similar target instances and relevant examples from the source domain. 

Moreover, to refine the local neighborhood and better control the influence of transferred source instances while mitigating the risk of negative transfer, we train a contrastive learning-based encoder $E$ using the unlabeled unified set \(\mathcal{I}^{x_t}\), following the SCARF method proposed by \cite{bahriscarf}. 
SCARF is an SSL method designed specifically for tabular data, which is characterized by heterogeneous and unordered features. In this work, SCARF is used to learn rich semantic embeddings from the unlabeled unified set by contrasting ordinal and corrupted versions of each instance. 
Briefly, SCARF creates corrupted views by replacing a random subset of features in each instance from \(\mathcal{I}^{x_t}\) with values sampled from their marginal distributions. 
The model is then trained to bring the representations of original and corrupted instances closer (positive pairs) while pushing apart the representations of different instances (negative pairs). The detailed process of the contrastive learning method is given in section \ref{SSLM}. 
The pretrained encoder $E$, obtained via this contrastive learning process, serves as the feature representation generator. 
Specifically, both $\bm{x}_{t}$ and the instances in \(\mathcal{I}^{x_t}\) are passed through $E$, which projects them into a shared latent space. We compute the latent embeddings of $\bm{x}_{t}$ and each instance in \(\mathcal{I}^{x_t}\), and measure their distances \(\bm{d}\) in the embedding space using a suitable distance metric depending on the specific context. 
To assign weights, we apply an exponential kernel over these distances, i.e., $ \bm{\pi}_{\bm{x}_t} = \exp (\frac{-\bm{d}^2}{2\sigma^2})$, where $\sigma$ is the width of the kernel. Instances closer to $\bm{x}_{t}$ in the latent space receive higher weights, while those farther away are assigned lower weights.  

Last, in \textbf{Step 3}, these combined weighted instances 
are passed through target black-box model $f_{target}$ to obtain the predictions. 
A linear surrogate model \( g_{\bm{x}_{t}} \) is then trained on these weighted instances and outputs, and the explanation produced by ITL-LIME can be formally defined as follows:
\begin{equation}
\xi(\bm{x_t}) = \arg\min_{g_{\bm{x}_{t}} \in G} \left[ \mathcal{L}(f_{target}, g_{\bm{x}_{t}}, \bm{\pi}_{\bm{x}_t}) + \Omega(g_{\bm{x}_{t}}) \right]
\label{eq:lime}
\end{equation}
where
\( \mathcal{L}(f_{target}, g_{\bm{x}_{t}}, \pi_{\bm{x}_t}) \) computes how well \( g_{\bm{x}_{t}} \) approximates \( f_{target} \) in the identified neighborhood of \( \bm{x_t} \), weighted by \( \bm{\pi}_{\bm{x}_t} \),
and \( \Omega(g_{\bm{x}_{t}}) \) controls the complexity of \( g_{\bm{x}_{t}} \), e.g., L$1$/L$2$ regularization. %
Algorithm~\ref{alg:proposed_method} outlines the ITL-LIME algorithm in detail.

\begin{algorithm} [h]
\caption{Proposed ITL-LIME Algorithm}
\label{alg:proposed_method}

\begin{flushleft}
\textbf{Input:} Target instance of interest $\bm{x}_{t}$ to be explained, trained black-box model $f_{target}$, source dataset \(\mathcal{D}_S\), target dataset \(\mathcal{D}_T\), number of source clusters $K$, transferred source sample to target neighborhood size $\xi$, encoder network $E$, and pre-train head network $h$. \\
\textbf{Output:}  Explanation model $g_{\bm{x}_{t}}$. \\
\textbf{Procedure:}
\end{flushleft}
\begin{algorithmic}[1]

\State Randomly select \(K\) data points as the initial centroid \( \mathcal{C} = \{\bm{c}_1, \bm{c}_2, \ldots, \bm{c}_K\} \subseteq \mathcal{D}_S \).
\State $\{I_1, I_2, \cdots, I_K\} \leftarrow$ Each data point $\bm{x}_s$ is assigned to the nearest centroid.
\State Update $\mathcal{C}$, Swap $\bm{c}_k$ with $\bm{x}_s$ which has the minimal sum of distances from each point:
$\displaystyle \underset{\bm{x}_s \in I_k}{\text{min}} \; \sum_{i=1}^n  \text{d}(\bm{x}_{s}, \bm{x}_i)$.
\State Repeat steps 2-3 until no further centroid swaps occur or the maximum number of iterations is reached.
\State \(\bm{c}_k^{x_t}\) $\leftarrow$ Calculate pairwise distance between $\bm{x}_{t}$ and each centroid in $\mathcal{C}$.
\State \(\mathcal{I}_{k}^{x_t}\) $\leftarrow$ Get the corresponding source real instances cluster set belongs to \(\bm{c}_k^{x_t}\).
\State \(\mathcal{I}_{tar}^{x_t}\) $\leftarrow$ Identify target neighborhood instances to $\bm{x}_t$ using $\xi$.
\State Unified neighborhood set by combining \(\mathcal{I}_{tar}^{x_t}\) and \(\mathcal{I}_{k}^{x_t}\) instances: \( \mathcal{I}^{x_t} \leftarrow \mathcal{I}_{k}^{x_t} \ \cup \ \mathcal{I}_{tar}^{x_t} \).

\State For each ${\bm{s}}^{(i)}\in \mathcal{I}^{x_t}$, create a corrupted view $\tilde{\bm{s}}^{(i)}$ by replacing a random subset of features.
\State Pass ${\bm{s}}^{(i)}$ and $\tilde{\bm{s}}^{(i)}$ through network $E$.
\State Generate embeddings ${z}^{(i)} \leftarrow h(E(\bm{s}^{(i)}))$ and $\tilde{z}^{(i)} \leftarrow h(E(\tilde{\bm{s}}^{(i)}))$ and compute pairwise similarity.
\State Update networks $E$ and $h$ to minimize loss in Eq. (\ref{eq:ssl_loss}) using SGD.
\State Calculate latent embeddings of $\bm{x}_{t}$ and each instance $\bm{s}^{(i)}$ in \(\mathcal{I}^{x_t}\), and measure their distances on the embedded space.
\State Calculate neighborhood instance weights using $ \bm{\pi}_{\bm{x}_t} \leftarrow \exp (\frac{-\bm{d}^2}{2\sigma^2})$.
\State Obtain predictions from black-box model $f_{target}$. 
\State Fit a linear interpretable model $g_{\bm{x}_{t}}$ using Eq. (\ref{eq:lime}).
\begin{flushleft}
\Statex \Return explanation model $g_{\bm{x}_{t}}$. 
\end{flushleft}
\end{algorithmic}
\end{algorithm}

\subsection{Self-supervised contrastive learning-based weighting method} \label{SSLM}
We adopt the SCARF method~\cite{bahriscarf} to learn meaningful representations from the unified source and target instance set \(\mathcal{I}^{x_t}\) and use the resulting encoder \(E\) as a similarity-based weighting function for better defining locality. The core idea is to train \(E\) to capture semantic patterns shared across both domains by generating two slightly perturbed views of the same instance and encouraging the model to recognize them as equivalent. 
The training procedure for \(E\) is outlined below.

Given the unified set \(\mathcal{I}^{x_t}\), for each mini-batch $N$ to create a corrupted version $\tilde{\bm{s}}^{(i)}$ for each instance \( \{ \bm{s}^{(i)} \}_{i=1}^N \subseteq \mathcal{I}^{x_t} \) using contrastive method, we randomly sample some fraction of the features uniformly and replace all of those features with the random value sample from the feature's empirical marginal distribution. Then we pass the original instance (${\bm{s}}^{(i)}$) and its corrupted version ($\tilde{\bm{s}}^{(i)}$) through encoder network $E$, which consists on four-layer neural network, whose output we pass through the pre-train head network $h$, to generates embeddings ${z}^{(i)} = h(E(\bm{s}^{(i)}))$ and $\tilde{z}^{(i)} = h(E(\tilde{\bm{s}}^{(i)}))$, respectively, which are $L_2$-normalized so that ${z}^{(i)}$ and $\tilde{z}^{(i)}$ lies on unit hyper sphere \cite{chen2020simple} and then compute the pairwise similarity as $ m_{i,j} = (z^{(i)})^T \tilde{z}^{(j)}$ ${/}$ $(\parallel {z}^{(i)}\parallel_2$ $\ast$ $\parallel \tilde{z}^{(j)} \parallel_2)$ for $i,j \in [N]$.
The InfoNCE contrastive loss ($\mathcal{L}_c$) function \cite{xu2023self} is adopted that encourage ${z}^{(i)}$ and $\tilde{z}^{(i)}$ to be close for all $i$ and ${z}^{(i)}$ and $\tilde{z}^{(j)}$ to be dissimilar for $i \neq j$, as shown below. 
\begin{equation}
\scalebox{0.8}{$
     \begin{aligned}[t] &
     \mathcal{L}_c = \frac{1}{N} \sum_{i=1}^N -\log (\frac{\exp\left(\frac{m_{i,i}}{\tau}\right)}{\frac{1}{N} \sum_{k=1}^N \exp\left(\frac{m_{i,j}}{\tau}\right)}) \end{aligned}
 $}
\label{eq:ssl_loss}
\end{equation} 
where $\tau$ is a hyperparameter that controls the sharpness of the softmax distribution over similarity scores. We used stochastic gradient descent (SGD) \cite{bottou2012stochastic} to tune the parameters of both $E$ and $h$. Once after training, we discard the pre-trained head network $h$ and use the encoder network $E$ as a weighting function.  

\section{Experiments} \label{EXP}
In this section, we first explain the experimental setup. Next, we design experiments to answer the following research questions (RQ):
\begin{itemize}
    \item RQ$1$: How does ITL-LIME improve local fidelity compared to standard and state-of-the-art (SOTA) methods in low-resource data settings?
    \item RQ$2$: To what extent does ITL-LIME improve the stability of explanations across the same instances and multiple runs?
    \item RQ$3$: How robust are ITL-LIME explanations to small perturbations in input data? 
    \item RQ$4$: How do ITL-LIME's components, such as source instance transfer and SSL-based weighting method, affect the overall local fidelity of the proposed model? 
    \item RQ$5$: How sensitive is the proposed ITL-LIME framework to key hyperparameters such as the number of source clusters $K$, and the transferred source sample to target neighborhood size ratio $\xi$?
\end{itemize}

\subsection{Experimental Setup}
\subsubsection{Datasets}
We conduct extensive experiments on two real-world healthcare datasets. The first dataset is the diabetes dataset \cite{kaggle_diabetes_dataset} that contains electronic health records (EHR) collected across the United States (U.S) states and territories between $2015$ and $2020$. It includes $10$ features containing demographic information and medical conditions. The output label is binary, where $1$ indicates diabetes and $0$ indicates no diabetes.
The dataset consists of three distinct subgroups that exhibit different feature distributions: one representing the mainland U.S, and two representing remote islands (the Virgin Islands in the Caribbean, and Guam in the Pacific Ocean). We designate the mainland data as the Diabetes Source Data (\textit{DSD}), the Virgin Islands as Diabetes Target Data $1$ (\textit{DTD1}), and Guam as Diabetes Target Data $2$ (\textit{DTD2}). 
\textit{DSD}, \textit{DTD1}, and \textit{DTD2} contain $96,738$, $763$, and $523$ patient records, respectively.

The other dataset is the student depression dataset \cite{OpenML2020StudentMentalHealth} that includes $16$ features, containing demographic information, academic metrics, and lifestyle features with binary output, where $0$ represents no depression and $1$ indicates students have depression. Based on the demographic feature, we define the Western India data as the Student Source Data (\textit{SSD}), and the East and South India data as Student Target Data $1$ (\textit{STD1}) and Student Target Data $2$ (\textit{STD2}), respectively, each with distinct feature distributions. 
The \textit{SSD} contains $11,440$ patient records, \textit{STD1} contains $650$ patient records and \textit{STD2} contains $625$ patient records.
\subsubsection{Implementation details}
Two representative black-box models, including Gaussian Support Vector Machine (G-LIBSVM) and Deep Neural Network (DNN), were trained on each target domain dataset for explanation. Hyperparameters were tuned using grid search with cross-validation. The DNN architecture included $4$ hidden layers with $256$, $128$, $64$, and $32$ neurons, respectively, and one output layer.
For ITL-LIME, we selected $K=15$ for $K$-medoid clustering from a predetermined range $\{11,\ 13,\ 15,\ 17,\ 19,\ 21,\ 23,\ 25\}$ in the source data based on silhouette analysis. We determine the transferred source sample size relative to the target neighborhood size $\xi$ based on the experimental trials from the values $\{1\!:\!0.5,\ 1\!:\!0.75,\ 1\!:\!1\}$.
For the contrastive learning-based encoder training, we adopted the default hyperparameters suggested in \cite{bahriscarf}, with batch size $= 64$, learning rate = $0.001$ with ADAM optimizer, $\tau=1.0$, the encoder ($E$) network contains $4$-layer ReLU MLP, hidden dimension $256$, pre-training head ($h$) with $2$-layer ReLU MLPs, and dropout rate $0.1$. We tuned the corruption rate parameters $c$ based on validation performance, within a suggested range of $c \in \{0.4,0.6,0.8\}$, selecting $c=0.6$ as they yielded the best validation accuracy in our setting.
We randomly selected $40$ instances from each target dataset for their local explanations. The final results reported were the average test metrics calculated across these $40$ corresponding explanation models.

\subsubsection{Baselines}
We compare ITL-LIME with standard and SOTA LIME-based methods, including Bay-LIME \cite{zhao2021baylime}, S-LIME \cite{zhou2021s}, US-LIME  \cite{saadatfar2024us}, A-LIME \cite{shankaranarayana2019alime}, D-LIME \cite{make3030027}, and LIME \cite{ribeiro2016should}, for each target dataset. 
Bay-LIME enhances explanation stability and fidelity by incorporating prior knowledge based on Bayesian reasoning \cite{zhao2021baylime}. 
S-LIME utilizes a hypothesis-testing framework based on the central limit theorem to determine the number of perturbed samples required for stable explanations \cite{zhou2021s}.
US-LIME utilizes uncertainty sampling and generates more focused perturbed samples close to the original instance and the black-box model decision boundary \cite{saadatfar2024us}. 
A-LIME utilizes an autoencoder as a weighting function to improve the locality and fidelity of explanations \cite{shankaranarayana2019alime}.
D-LIME builds clusters to access a subset of real instances to build a local interpretable model to improve fidelity and stability \cite{make3030027}.  

\subsubsection{Explanation evaluation metrics}
We used widely adopted explanation metrics to evaluate the quality and reliability of model explanation, including:

\textbf{($1$) Fidelity:} It measures how well the local surrogate model approximates the behavior of the global black-box model locally. Specifically, it compares the predictions made by the black-box model with the surrogate model.  
Since the adopted datasets are for classification tasks, we evaluate the local fidelity using the F1-score and the Area Under the ROC Curve (AUC). High fidelity indicates that the explanation is accurate and faithful to the model's actual behavior. 

\textbf{($2$) Stability:} It evaluates the consistency of the explanation across multiple runs. To compute stability and consistency of explanation, we adapted the Jaccard Coefficient (JC) that assesses the consistency of feature importance across the same input observations and multiple runs. It computes the similarity between two sets of top $k$ features using: 
\begin{equation}
Jacard(A, B) = \frac{|A \cap B|}{|A \cup B|} 
\end{equation} \label{eq:jc}
where $A$ and $B$ are sets of top $k$ features provided by the explanation model. $Jaccard (A, B) = 1$ indicates that $A$ and $B$ are identical, indicating stable explanations. $Jaccard(A, B) = 0$ indicates that $A$ and $B$ share no overlap ($|A \cap B| = 0$), indicating instability. We set $k=3$ for the experiments to evaluate whether the top features identified by the explanation method remain consistent across multiple runs. We generate explanations around the target instance for the top ($k=3$) features and repeat this process over five iterations.

\textbf{($3$) Robustness:} We adapted the Local Lipschitz Estimator (LLE) \cite{datta2021machine} to compute robustness of the explanation models, which measures the sensitivity of an explanation to small changes in the input observations. It is calculated as:
\begin{equation}
L(\bm{x}) = \max_{\bm{x}^{'} \in B_\epsilon(\bm{x})} \left( 
\frac{\|f(\bm{x}) - f(\bm{x}^{'}) \|}{\| \bm{x} - \bm{x}^{'} \|} 
\right) 
\end{equation} \label{eq:lle}
where $f(\cdot)$ indicates feature importance scores provided by the interpretable model, and $B_\epsilon(\bm{x})$ is the neighborhood of radius $\epsilon$ centered at $\bm{x}$. Higher values $(L(\bm{x}) \gg 1)$ indicate high sensitivity where small changes significantly alter the explanation, indicating the model is less robust to input change. A lower LLE value of ($0$ < $LLE(\bm{x})$ $\ll$ $1$) indicates high robustness and low sensitivity to input changes \cite{jordan2020exactly}. 
We added small Gaussian noise (ranging from $0.01$ to $0.1$) to the continuous features, while for the categorical features, we randomly sampled values for each input sample to simulate noise and generated the explanations, and computed LLE over five iterations to report the average score. Lower LLE values indicate more robust and stable explanations. 

\subsection{Fidelity (RQ$1$)}
We conducted extensive experiments to demonstrate the performance of the proposed ITL-LIME in improving the fidelity of the explanation and compared its performance with baseline methods. Further, we compared the predictive performance of explanation models with the true labels on all real test instances.

\subsubsection{Local fidelity}
The average testing results of all the models' local fidelity are presented in Table \ref{tab:local_fidelity}; the bold numbers indicate the best results for each metric. ITL-LIME and other baseline models are evaluated on $40$ randomly selected instances from the target sets for comparison purposes.
As shown in Table \ref{tab:local_fidelity}, ITL-LIME consistently achieved the highest F1-score and AUC across all black-box models and target sets. 
ITL-LIME, compared to standard LIME, shows a significant improvement in both metrics. 
For instance, \textit{DTD1}-G-LIBSVM settings, ITL-LIME outperforms LIME by $8.6\%$ points in F1-score ($0.6917$ vs. $0.6061$) and AUC with $9.6\%$ points ($0.9029$ vs. $0.8069$).
Similar trends were observed on \textit{DTD2}, where ITL-LIME improves F1-score from $0.5636$ to $0.6615$ and AUC from $0.8041$ to $0.9071$, when using DNN.
On \textit{STD1} and \textit{STD2}, ITL-LIME also maintained its advantage while achieving the highest F1-score ($0.8809$) and AUC ($0.8892$) with \textit{STD1}-G-LIBSVM, outperforming all other models, including the strongest baselines D-LIME (F1 = $0.8204$, AUC = $0.8574$). Similarly, on \textit{STD2}-DNN, ITL-LIME achieved the highest F1-score ($0.8731$) and AUC ($0.8982$) compared to all baseline models.
Importantly, while several baselines like S-LIME, US-LIME, and A-LIME improve upon standard LIME, they underperform compared to ITL-LIME. For example, D-LIME, which also leverages real instances, fails to match ITL-LIME’s performance, likely because it lacks ITL-LIME’s instance transfer mechanism and contrastive encoder weighting, which enhance the local approximation of the black-box decision boundary.

\begin{table*}[ht]
\caption{The average testing fidelity results for all explanation methods on $40$ randomly selected test instances from the target domain. Higher values ($\uparrow$) indicate better fidelity.}
\label{tab:local_fidelity}
\centering
\renewcommand{\arraystretch}{0.95}
\resizebox{\textwidth}{!}{
\begin{tabular}{l l
    *{14}{c}}
\toprule
\textbf{Datasets} & \textbf{Black-box}
& \multicolumn{2}{c}{\textbf{Bay-LIME}} 
& \multicolumn{2}{c}{\textbf{S-LIME}} 
& \multicolumn{2}{c}{\textbf{US-LIME}} 
& \multicolumn{2}{c}{\textbf{A-LIME}} 
& \multicolumn{2}{c}{\textbf{D-LIME}} 
& \multicolumn{2}{c}{\textbf{LIME}} 
& \multicolumn{2}{c}{\textbf{ITL-LIME}} \\
\cmidrule(lr){3-4} \cmidrule(lr){5-6} \cmidrule(lr){7-8} \cmidrule(lr){9-10} \cmidrule(lr){11-12} \cmidrule(lr){13-14} \cmidrule(lr){15-16} 
& & \textbf{F1-Score} & \textbf{AUC}
  & \textbf{F1-Score} & \textbf{AUC}
  & \textbf{F1-Score} & \textbf{AUC}
  & \textbf{F1-Score} & \textbf{AUC}
  & \textbf{F1-Score} & \textbf{AUC}
  & \textbf{F1-Score} & \textbf{AUC}
  & \textbf{F1-Score} & \textbf{AUC} \\
\midrule
\multirow{2}{*}{\textit{DTD1}} & G-LIBSVM 
  & 0.6299 & 0.8326 & 0.6353 & 0.8313 & 0.6449 & 0.8560 & 0.6513 & 0.8767 & 0.6610 & 0.8762 & 0.6061 & 0.8069 & \textbf{0.6917} & \textbf{0.9029} \\
& DNN 
  & 0.5960 & 0.8058 & 0.6413 & 0.8591 & 0.6644 & 0.8694 & 0.6346 & 0.8852 & 0.6667 & 0.8639 & 0.5727 & 0.8092 & \textbf{0.6816} & \textbf{0.9038} \\
\midrule

\multirow{2}{*}{\textit{DTD2}} & G-LIBSVM 
  & 0.6371 & 0.8349 & 0.6368 & 0.8377 & 0.6370 & 0.8623 & 0.6371 & 0.8877 & 0.6478 & 0.8604 & 0.5759 & 0.8049 & \textbf{0.6961} & \textbf{0.9119} \\
& DNN 
  & 0.6524 & 0.8835 & 0.6328 & 0.8459 & 0.6456 & 0.8615 & 0.6417 & 0.8890 & 0.6343 & 0.8524 & 0.5636 & 0.8041 & \textbf{0.6615} & \textbf{0.9071} \\
\midrule

\multirow{2}{*}{\textit{STD1}} & G-LIBSVM 
  & 0.8219 & 0.8537 & 0.7884 & 0.8130 & 0.8355 & 0.8571 & 0.8236 & 0.8547 & 0.8204 & 0.8574 & 0.7434 & 0.7500 & \textbf{0.8809} & \textbf{0.8892} \\
& DNN 
  & 0.8332 & 0.8662 & 0.8057 & 0.8379 & 0.8276 & 0.8528 & 0.8366 & 0.8667 & 0.8212 & 0.8564 & 0.7509 & 0.7614 & \textbf{0.8963} & \textbf{0.9180} \\
\midrule

\multirow{2}{*}{\textit{STD2}} & G-LIBSVM 
  & 0.8100 & 0.8505 & 0.8289 & 0.8598 & 0.8279 & 0.8502 & 0.8358 & 0.8707 & 0.8476 & 0.8718 & 0.8055 & 0.8088 & \textbf{0.8936} & \textbf{0.9061} \\
& DNN 
  & 0.8155 & 0.8430 & 0.8170 & 0.8305 & 0.8243 & 0.8425 & 0.8312 & 0.8497 & 0.8367 & 0.8409 & 0.7690 & 0.7717 & \textbf{0.8731} & \textbf{0.8982} \\
\bottomrule
\end{tabular}
}
\end{table*}

\subsubsection{Evaluation against true labels}
Following \cite{NEURIPS2020_ce758408}, we also compared the predictive performance of all explanation models on real test instances with true labels. The results are given in Table \ref{tab:true_fidelity}, the bold numbers indicate the best result for each metric. 
As expected, ITL-LIME's performance on real instances aligned more closely with true labels compared with LIME without transfer. For instance, on the \textit{DTD1}-G-LIBSVM, ITL-LIME outperforms LIME by $9.5$\% points in F1-score ($0.5134$ vs. $0.4180$) and $6.3$ points in AUC ($0.8771$ vs. $0.8140$). Similar trends were observed for all other datasets.
It is likely due to ITL-LIME utilizing real instances during explanation that more closely reflect the behavior of the black-box model, resulting in explanations that more accurately align with the true labels. 
ITL-LIME also outperformed other baseline models, including Bay-LIME, S-LIME, US-LIME, A-LIME, and D-LIME across all black-box models for all target sets.
These results demonstrate that ITL-LIME not only approximates the decision-making process of a black-box model accurately but also maintains a stronger alignment with real-world predictions. 
\begin{table*}[ht]
\captionsetup{justification=raggedright,singlelinecheck=false}
\caption{The average performance results of all LIME-based explanation methods for real test instances.}
\label{tab:true_fidelity}
\centering
\renewcommand{\arraystretch}{0.95}
\resizebox{\textwidth}{!}{
\begin{tabular}{l l
    *{14}{c}}
\toprule
\textbf{Datasets} & \textbf{Black-box}
& \multicolumn{2}{c}{\textbf{Bay-LIME}} 
& \multicolumn{2}{c}{\textbf{S-LIME}} 
& \multicolumn{2}{c}{\textbf{US-LIME}} 
& \multicolumn{2}{c}{\textbf{A-LIME}} 
& \multicolumn{2}{c}{\textbf{D-LIME}} 
& \multicolumn{2}{c}{\textbf{LIME}} 
& \multicolumn{2}{c}{\textbf{ITL-LIME}} \\
\cmidrule(lr){3-4} \cmidrule(lr){5-6} \cmidrule(lr){7-8} \cmidrule(lr){9-10} \cmidrule(lr){11-12} \cmidrule(lr){13-14} \cmidrule(lr){15-16} 
& & \textbf{F1-Score} & \textbf{AUC}
  & \textbf{F1-Score} & \textbf{AUC}
  & \textbf{F1-Score} & \textbf{AUC}
  & \textbf{F1-Score} & \textbf{AUC}
  & \textbf{F1-Score} & \textbf{AUC}
  & \textbf{F1-Score} & \textbf{AUC}
  & \textbf{F1-Score} & \textbf{AUC} \\
\midrule
\multirow{2}{*}{\textit{DTD1}} & G-LIBSVM 
  & 0.4665 & 0.8053 
  & 0.4414 & 0.8260
  & 0.4507 & 0.8114
  & 0.4282 & 0.8338
  & 0.4659 & 0.8598 
  & 0.4180 & 0.8140
  & \textbf{0.5134} & \textbf{0.8771} \\
& DNN 
  & 0.4759 & 0.8397 
  & 0.4679 & 0.8289 
  & 0.4681 & 0.8210 
  & 0.4753 & 0.8239  
  & 0.4748 & 0.8426
  & 0.4545 & 0.8155 
  & \textbf{0.5104} & \textbf{0.8875} \\
\midrule

\multirow{2}{*}{\textit{DTD2}} & G-LIBSVM 
  & 0.4676 & 0.8391 
  & 0.4788 & 0.8266 
  & 0.4735 & 0.8462 
  & 0.4570 & 0.8232 
  & 0.4737 & 0.8500 
  & 0.4045 & 0.8081 
  & \textbf{0.5078} & \textbf{0.8815} \\
& DNN 
  & 0.4765 & 0.8481 
  & 0.4685 & 0.8421 
  & 0.4666 & 0.8460 
  & 0.4725 & 0.8424 
  & 0.4571 & 0.8416 
  & 0.4778 & 0.8063 
  & \textbf{0.5029} & \textbf{0.8693} \\
\midrule

\multirow{2}{*}{\textit{STD1}} & G-LIBSVM 
  & 0.7121 & 0.7487 
  & 0.7209 & 0.7499 
  & 0.7098 & 0.7395 
  & 0.7165 & 0.7451 
  & 0.7219 & 0.7546 
  & 0.6837 & 0.7239 
  & \textbf{0.7589} & \textbf{0.7822} \\
& DNN 
  & 0.7164 & 0.7475 
  & 0.7141 & 0.7438 
  & 0.7195 & 0.7423 
  & 0.7130 & 0.7398 
  & 0.7249 & 0.7462 
  & 0.6670 & 0.7058 
  & \textbf{0.7421} & \textbf{0.7704} \\
\midrule

\multirow{2}{*}{\textit{STD2}} & G-LIBSVM 
  & 0.7195 & 0.7362 
  & 0.7159 & 0.7434 
  & 0.7237 & 0.7304 
  & 0.7108 & 0.7408 
  & 0.7157 & 0.7458 
  & 0.6993 & 0.7061 
  & \textbf{0.7637} & \textbf{0.7849} \\
& DNN 
  & 0.7087 & 0.7452 
  & 0.7140 & 0.7377 
  & 0.7044 & 0.7262 
  & 0.7122 & 0.7457 
  & 0.7121 & 0.7518 
  & 0.6539 & 0.6909 
  & \textbf{0.7490} & \textbf{0.7720} \\
\bottomrule
\end{tabular}
}
\end{table*}
\subsection{Stability (RQ$2$)} 
Table \ref{tab:stability_results} presents the average stability results of all explanation models across black-box models; the bold numbers indicate the best results. 
ITL-LIME consistently achieved the highest JC of $100$\% across all datasets and black-box models, outperforming all baseline models, indicating more stable and consistent selection of features for the same input instance across multiple runs.
For instance, \textit{DTD1}-DNN setting, ITL-LIME obtained $100$\% JC, compared to $0.7895$ for standard LIME, $0.9125$ for US-LIME, and $0.9712$ for S-LIME. Similarly with \textit{DTD2}-G-LIBSVM, ITL-LIME achieved $100$\%, while Bay-LIME reached $0.9700$, S-LIME achieved $0.9487$, US-LIME achieved $0.9450$, and A-LIME achieved $0.8712$, indicating these methods utilized perturbed samples, where new instances were generated by random sampling, which may lead to inconsistent and unreliable explanations.
The same trends were observed with \textit{STD1} and \textit{STD2} datasets, where ITL-LIME attained $100$\% JC, compared to $0.8695$ for LIME, $0.9380$ for A-LIME, $0.9762$ for US-LIME, $0.9875$ for S-LIME, and $100$\% for Bay-LIME with \textit{STD1}-G-LIBSVM settings.
These improvements are likely due to ITL-LIME, instead of random perturbation, utilizing real source instances combined with neighboring target real instances that more stably represent the true local behavior of the black-box model, resulting in more stable and reliable explanations. 
Additionally, the improvement in JC can be attributed to the contrastive learning based encoder’s ability to prioritize relevant instances by assigning higher weights, thus reducing the influence of less relevant instances with lower weights. 
Among the baselines, D-LIME also obtained $100$\% JC, as it also utilizes real instances for explanation. However, it lacks access to source domain information and contrastive weighting, leading to lower fidelity than ITL-LIME. On the other hand, ITL-LIME not only improves stability but also enhances locality and fidelity.

\begin{table*}[ht]
\footnotesize 
\caption{The average stability results for all explanation methods on $40$ randomly selected test instances from the target domain. Higher values ($\uparrow$) indicate better stability.}
\label{tab:stability_results}
\centering
\renewcommand{\arraystretch}{0.80}
\begin{tabular}{p{1.9cm} p{2cm} p{1.5cm} p{1.5cm} p{1.5cm} p{1.5cm} p{1.5cm} p{1.5cm} p{1.5cm}}
\toprule
\textbf{Datasets} & \textbf{Black-box} 
& \textbf{Bay-LIME} 
& \textbf{S-LIME} 
& \textbf{US-LIME} 
& \textbf{A-LIME} 
& \textbf{D-LIME} 
& \textbf{LIME} 
& \textbf{ITL-LIME} \\
\midrule

\multirow{2}{*}{\textit{DTD1}} & G-LIBSVM 
  & 0.9538 & 0.9700 & 0.9641 & 0.9187 & 1.0000 & 0.8888 & \textbf{1.0000} \\
& DNN 
  & 1.0000 & 0.9712 & 0.9125 & 0.8555 & 1.0000 & 0.7895 & \textbf{1.0000} \\
\midrule

\multirow{2}{*}{\textit{DTD2}} & G-LIBSVM 
  & 0.9700 & 0.9487 & 0.9450 & 0.8712 & 1.0000 & 0.7956 & \textbf{1.0000} \\
& DNN 
  & 1.0000 & 0.9350 & 0.9350 & 0.9418 & 1.0000 & 0.7870 & \textbf{1.0000} \\
\midrule

\multirow{2}{*}{\textit{STD1}} & G-LIBSVM 
  & 1.0000 & 0.9875 & 0.9762 & 0.9380 & 1.0000 & 0.8695 & \textbf{1.0000} \\
& DNN 
  & 1.0000 & 0.9862 & 1.0000 & 0.9349 & 1.0000 & 0.8283 & \textbf{1.0000} \\
\midrule

\multirow{2}{*}{\textit{STD2}} & G-LIBSVM 
  & 1.0000 & 0.9845 & 0.9850 & 0.9850 & 1.0000 & 0.9166 & \textbf{1.0000} \\
& DNN 
  & 1.0000 & 1.0000 & 1.0000 & 0.9125 & 1.0000 & 0.8633 & \textbf{1.0000} \\
\bottomrule
\end{tabular}
\end{table*}

\subsection{Robustness (RQ$3$)}
Table \ref{tab:robustness_results} presents the average robustness results of all explanation models across black-box models; the bold numbers indicate the best results. 
Across all datasets and black-box models, ITL-LIME consistently achieved the lowest LLE, outperforming all baseline methods. For example, in the \textit{DTD1}-G-LIBSVM setting, ITL-LIME attained an LLE of $0.7693$, compared to $1.1102$ for LIME, $0.8525$ for D-LIME, and $0.9307$ for Bay-LIME. Similarly, on the \textit{STD1}-DNN settings, ITL-LIME achieved $0.7949$, while LIME reached $1.0106$, indicating significantly higher sensitivity to input changes. With \textit{STD2}-G-LIBSVM settings, ITL-LIME achieved an LLE of $0.7054$, compared to $0.9570$ for LIME, $0.7524$ for D-LIME, $0.7767$ for A-LIME,
 indicating high sensitivity to input noise. 
This improvement stems from ITL-LIME’s use of real source and target instances, instead of randomly perturbed samples, to construct the local neighborhood for explanation. Additionally, the contrastive encoder assigns higher weights to semantically relevant instances, reducing the influence of noisy or less informative examples. As a result, ITL-LIME generates explanations that are less sensitive to small input variations, making them more robust and reliable. 
Among the baselines, D-LIME exhibits moderate robustness, due to its use of real instances. However, lacking access to source domain information and contrastive weighting, it fails to match the performance of ITL-LIME. 
Overall, these results highlight the effectiveness of ITL-LIME in handling noise and are less sensitive to input change, which can be favorable, especially in sensitive domains such as clinical decision support explanations.

\begin{table*}[ht]
\footnotesize 
\caption{The average robustness results for all explanation methods on $40$ randomly selected test instances from the target domain. Lower values ($\downarrow$) indicate better robustness.}
\label{tab:robustness_results}
\centering
\renewcommand{\arraystretch}{0.80}
\begin{tabular}{p{1.9cm} p{2cm} p{1.5cm} p{1.5cm} p{1.5cm} p{1.5cm} p{1.5cm} p{1.5cm} p{1.5cm}}
\toprule
\textbf{Datasets} & \textbf{Black-box} 
& \textbf{Bay-LIME} 
& \textbf{S-LIME} 
& \textbf{US-LIME} 
& \textbf{A-LIME} 
& \textbf{D-LIME} 
& \textbf{LIME} 
& \textbf{ITL-LIME} \\
\midrule

\multirow{2}{*}{\textit{DTD1}} & G-LIBSVM 
  & 0.9307 & 0.9285 & 0.8933 & 0.9079 & 0.8525 & 1.1102 & \textbf{0.7693} \\
& DNN 
  & 0.9054 & 0.8560 & 0.9274 & 0.9158 & 0.9223 & 1.0648 & \textbf{0.7891} \\
\midrule

\multirow{2}{*}{\textit{DTD2}} & G-LIBSVM 
  & 0.9228 & 0.8812 & 0.9670 & 0.9117 & 0.9023 & 1.2754 & \textbf{0.8151} \\
& DNN 
  & 0.9324 & 0.9409 & 0.9635 & 0.9486 & 0.8889 & 1.1882 & \textbf{0.8163} \\
\midrule

\multirow{2}{*}{\textit{STD1}} & G-LIBSVM 
  & 0.8360 & 0.8492 & 0.8575 & 0.8467 & 0.8381 & 0.8761 & \textbf{0.8173} \\
& DNN 
  & 0.8163 & 0.8618 & 0.8414 & 0.8623 & 0.8477 & 1.0106 & \textbf{0.7949} \\
\midrule

\multirow{2}{*}{\textit{STD2}} & G-LIBSVM 
  & 0.7609 & 0.7952 & 0.7843 & 0.7767 & 0.7524 & 0.9570 & \textbf{0.7054} \\
& DNN 
  & 0.8047 & 0.7889 & 0.8160 & 0.7949 & 0.7999 & 0.9888 & \textbf{0.7528} \\
\bottomrule
\end{tabular}
\end{table*}

\begin{table*}[ht]
\caption{Average fidelity results of ablation study for $40$ randomly selected test instances from the target domain.}
\captionsetup{justification=raggedright,singlelinecheck=false}
\label{tab:ablation_study}
\renewcommand{\arraystretch}{0.65}
\resizebox{\textwidth}{!}{
\begin{tabular}{l l
    r@{\,±\,}l r@{\,±\,}l
    r@{\,±\,}l r@{\,±\,}l
    r@{\,±\,}l r@{\,±\,}l
}
\toprule
\textbf{Datasets} & \textbf{Black-box}
& \multicolumn{4}{c}{\textbf{ITL-LIME w/o Encoder based Weighting}}
& \multicolumn{4}{c}{\textbf{ITL-LIME w/o Source ITL}}
& \multicolumn{4}{c}{\textbf{ITL-LIME}} \\
\cmidrule(lr){3-6} \cmidrule(lr){7-10} \cmidrule(lr){11-14}
& & \multicolumn{2}{c}{\textbf{F1-Score}} & \multicolumn{2}{c}{\textbf{AUC}}
  & \multicolumn{2}{c}{\textbf{F1-Score}} & \multicolumn{2}{c}{\textbf{AUC}}
  & \multicolumn{2}{c}{\textbf{F1-Score}} & \multicolumn{2}{c}{\textbf{AUC}} \\
\midrule
\textit{DTD2} & G-LIBSVM & \multicolumn{2}{c}{0.6690 $\pm$ 0.0807} & \multicolumn{2}{c}{0.8589 $\pm$ 0.0514 } & \multicolumn{2}{c}{0.6416 $\pm$ 0.0578} & \multicolumn{2}{c}{0.8456 $\pm$ 0.0364} & \multicolumn{2}{c}{\textbf{0.6961 $\pm$ 0.0726}} & \multicolumn{2}{c}{\textbf{0.9119 $\pm$ 0.0403}} \\
\textit{DTD2} & DNN & \multicolumn{2}{c}{0.6268 $\pm$ 0.0687} & \multicolumn{2}{c}{0.8624 $\pm$ 0.0457} & \multicolumn{2}{c}{0.5891 $\pm$ 0.0293} & \multicolumn{2}{c}{0.8438 $\pm$ 0.0099} & \multicolumn{2}{c}{\textbf{0.6615 $\pm$ 0.0750}} & \multicolumn{2}{c}{\textbf{0.9071 $\pm$ 0.0536}} \\
\midrule
\textit{STD1} & G-LIBSVM & \multicolumn{2}{c}{0.8454 $\pm$ 0.1088} & \multicolumn{2}{c}{0.8575 $\pm$ 0.0545} & \multicolumn{2}{c}{0.8109 $\pm$ 0.0649} & \multicolumn{2}{c}{0.8424 $\pm$ 0.0579} & \multicolumn{2}{c}{\textbf{0.8809 $\pm$ 0.0200}} & \multicolumn{2}{c}{\textbf{0.8892 $\pm$ 0.0286}} \\
\textit{STD1} & DNN & \multicolumn{2}{c}{0.8247 $\pm$ 0.0393} & \multicolumn{2}{c}{0.8448 $\pm$ 0.0390} & \multicolumn{2}{c}{0.8076 $\pm$ 0.0443} & \multicolumn{2}{c}{0.8452 $\pm$ 0.0426} & \multicolumn{2}{c}{\textbf{0.8963 $\pm$ 0.0204}} & \multicolumn{2}{c}{\textbf{0.9180 $\pm$ 0.0220}} \\
\bottomrule
\end{tabular}
}
\end{table*}
\subsection{Ablation Experiments (RQ$4$)}
To verify the impact of key components of ITL-LIME on the explanation performance, we design the following ablation experiments: 

($1$) Remove the contrastive learning encoder-based weighting and apply the default LIME weighting method for the proximity task. 

($2$) Remove instance-based transfer learning so no source domain is involved, and rely solely on the target domain, and weight each instance using the contrastive learning encoder.

The average fidelity results of the ablation study computed on $40$ test instances across all black-box models using \textit{DTD2} and \textit{STD1} are presented in Table \ref{tab:ablation_study}.
Our observations are as follows:
\begin{itemize} [noitemsep, topsep=0pt]
    \item A decrease in fidelity was observed when we applied the default LIME weighting instead of contrastive learning based encoder weighting for the proximity task. For instance, with \textit{DTD2}-G-LIBSVM settings, the ITL-LIME w/o encoder based weighting, F1-score decreased from $0.6961$ to $0.6690$, and AUC from $0.9119$ to $0.8589$, and with \textit{STD1}-DNN settings, F1-score dropped from $0.8963$ to $0.8247$, and AUC from $0.9180$ to $0.8448$. 
    This advantage arises because the contrastive learning encoder captures more robust and semantically aligned representations across both source and target domains, despite their distribution differences. It better preserves transferable patterns and is more resilient to noisy features. In contrast, the default LIME weighting, which relies on the original input space, can be sensitive to domain shifts and domain-specific noise, resulting in less reliable proximity-based weightings.
    \item When the source ITL was removed, but the encoder weighting was still applied, the performance dropped more significantly. For instance, with \textit{DTD2}-DNN settings, ITL-LIME w/o source ITL F1-score dropped from $0.6615$ to $0.5891$, and AUC from $0.9071$ to $0.8438$, and with \textit{STD1}-G-LIBSVM setup, F1-score decreased from $0.8809$ to $0.8109$, and AUC from $0.8892$ to $0.8424$. These results highlight that the absence of source ITL limits the performance of the model, leading to a drop in fidelity. Whereas ITL-LIME can improve the local explanation quality on small datasets by leveraging the carefully selected relevant source instances with encoder-based weightings from a data-rich source domain. 
\end{itemize}
\subsection{Sensitivity Analysis (RQ$5$)}
\subsubsection{Impact of source cluster count \(K\) on ITL-LIME's local fidelity}
We conducted a sensitivity analysis by varying the number of clusters \(K\) in the \(K\)-medoids clustering of the source domain, as \(K\) directly impacts cluster quality and the relevance of retrieved instances for explanation. For each value of $K$, we measured the average fidelity of ITL-LIME explanation on $40$ randomly selected instances from the \textit{STD1} dataset. 

\begin{figure}[ht]
    \centering
    \includegraphics[width=0.39\textwidth]{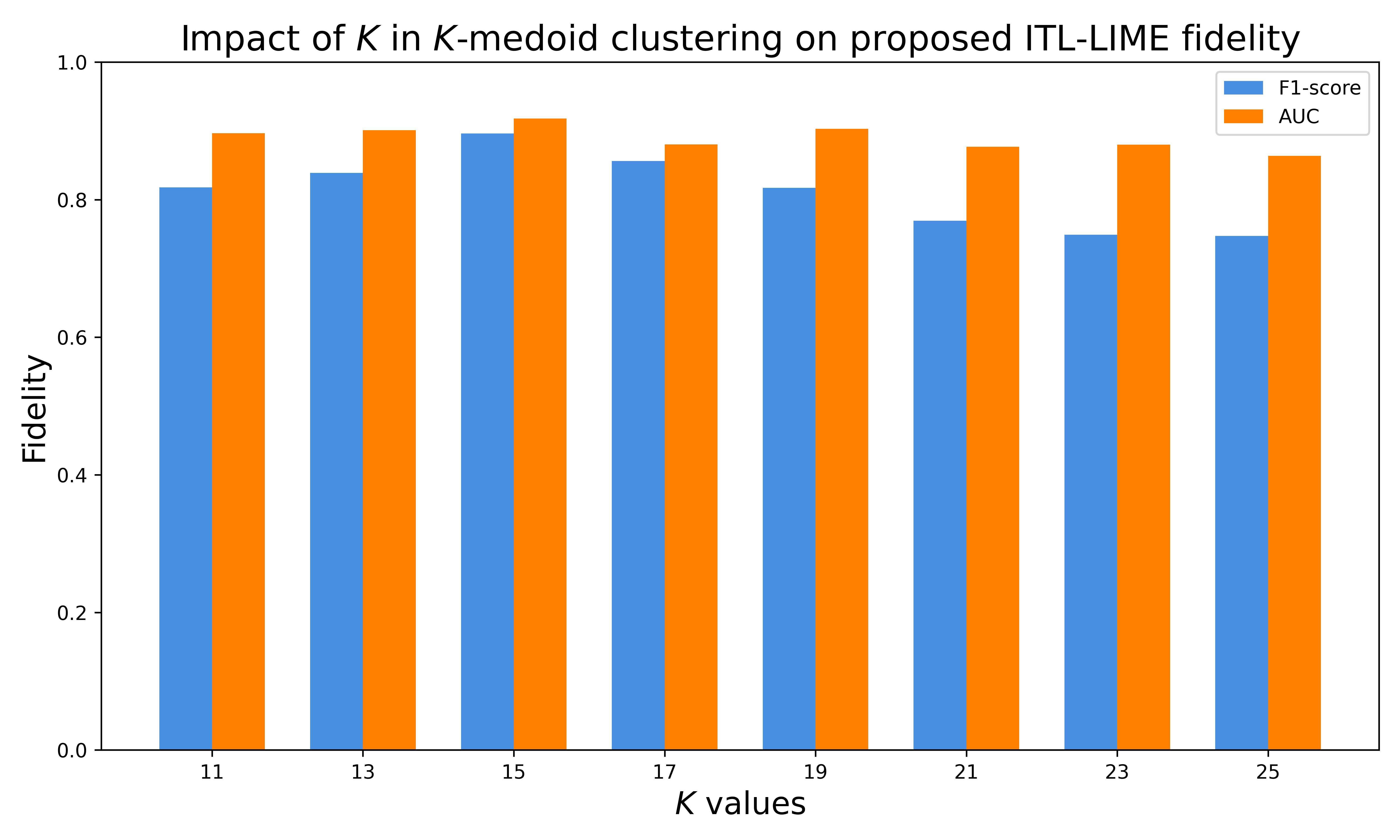}
    \caption{Impact of \(K\) on ITL-LIME fidelity across a DNN, evaluated on $40$ selected test instances from \textit{STD1}. }
    \label{fig:sensitivity_k_value}
\end{figure}

As shown in Figure \ref{fig:sensitivity_k_value}, ITL-LIME achieved consistently high fidelity for \(K\) between $11$ to $15$. For instance, ITL-LIME achieved an F1-score of $0.8179$, and an AUC of $0.8967$ at $K=11$, and with the best F1-score ($0.8963$) and AUC ($0.9180$) at $K=15$, which also corresponds to the optimal value tuned on the source domain.  
Beyond \(K=19\), the fidelity F1-score declined, i.e., at $K=21$, ITL-LIME achieved F1-score of $0.7694$, and declined to $0.7474$ at $K=25$, which can be attributed to excessive clustering leading to smaller clusters, thereby affecting the relevance of source instances for transfer. 
Overall, ITL-LIME demonstrates reasonable robustness to variations in \(K\). However, selecting an appropriate value of \(K\) on the source domain using techniques such as elbow or silhouette analysis is important to maintain high fidelity.
\subsubsection{Impact of transferred source sample vs. target neighborhood sample size ratio ($\xi$) on ITL-LIME Fidelity}
\begin{figure}[ht]
    \centering
    \includegraphics[width=0.4\textwidth]{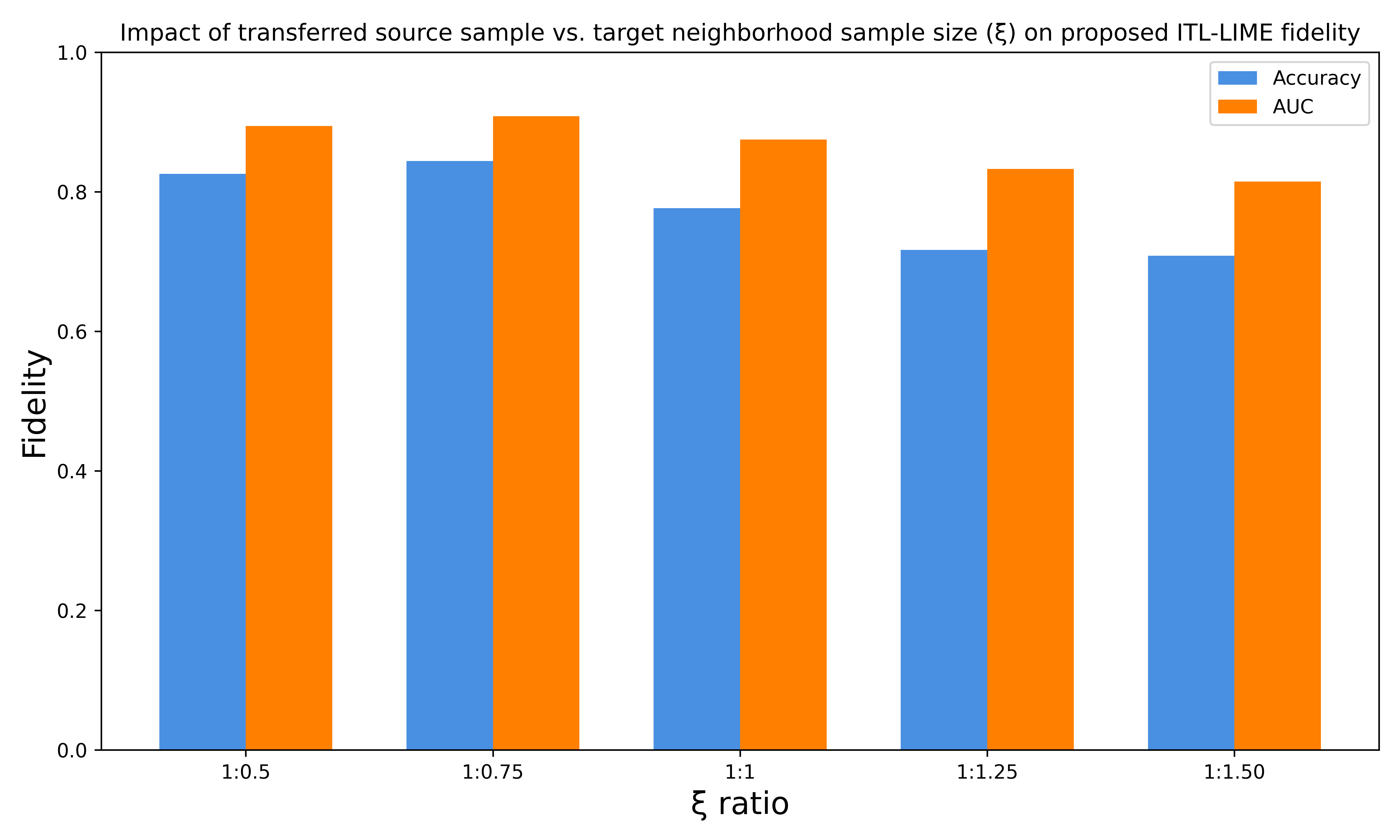}
    \caption{Impact of $\xi$ ratio on ITL-LIME fidelity across a DNN, evaluated on $40$ selected test instances from \textit{STD1}.}
    \label{fig:senstivity_ratio}
\end{figure}

We also conducted a sensitivity analysis by varying $\xi$ to examine its impact on ITL-LIME's fidelity performance.  
For each selected ratio, we measured the average fidelity of ITL-LIME explanation on $40$ randomly selected instances from the \textit{STD1} dataset. 
As shown in Figure \ref{fig:senstivity_ratio}, with lower $\xi$, such as $1\!:0.5$ or $1\!:0.75$, the model demonstrates higher fidelity. As $\xi$ increased from $1\!: 0.75$, we observed a decline in fidelity. For instance, with $\xi$ ratio $1\!:0.5$, ITL-LIME achieved F1-score of $0.8257$ and AUC of $0.8942$, while with a significant increase in $\xi$ ratio to $1\!:1.5$, F1-score dropped to $0.7084$ and AUC $0.8147$. 
 These results highlight that an excessive amount of target neighboring instances may adversely affect the ITL-LIME performance, resulting in less faithful explanations. This can be due to a larger target neighborhood size may result in less local capability, especially in small dataset settings. From our experimental trials, it is suggested to decide the $\xi$ ratio from the range $\{1\!:\!0.5,\ 1\!:\!0.75,\ 1\!:\!1\}$. 
\section{Conclusion} \label{CF}
In this study, we proposed a novel instance-based transfer learning LIME framework, ITL-LIME, that enhances the fidelity and stability of LIME explanation by transferring the real instances from the source domain to the data-scarce target domain. Instead of a random perturbation, ITL-LIME utilizes clustering methods to select source real instances and combine them with neighboring target instances to generate the local explanation. The contrastive learning-based encoder learns the semantic representation from source and target real instances and defines compact locality by assigning weights to each instance based on proximity to the target instance. 
By incorporating carefully selected and weighted relevant real instances from both source and target domains, ITL-LIME approximates the local behavior of the black-box model more accurately, resulting in improved fidelity and stability. 
Our extensive experiments on real-world datasets demonstrate that ITL-LIME successfully enhances the local fidelity as well as stability in the limited-data domain compared to other baseline models. 
In the future, we intend to investigate the performance impact across a more extensive and diverse domain datasets to further enhance real-world applicability. Furthermore, an analysis of computational complexity and runtime overhead will be conducted to evaluate the method’s practical feasibility.
\begin{acks}
R.R. is funded by MU MIPS scholarship; G.W. is supported by the Google Academic Research Award ’24; M. F. is supported by the EU project RECITALS (GA 101168490) and the German BMBF project FEDCOV (GA 100685937).
\end{acks}

\smallskip
\smallskip
\noindent 
\textbf{GenAI Usage disclosure statement}

This paper utilized Generative AI tools in the following stages of the research process:
\begin{itemize}
    \item Writing Assistance: ChatGPT (GPT-4) was used to assist in improving sentence structure, enhancing clarity and flow, and checking grammar.
    \item Code Debugging: ChatGPT (GPT-4) was employed to assist in debugging Python code, particularly in areas where errors were encountered during the experiments.
\end{itemize}
All AI-generated content was manually reviewed for accuracy and relevance. The use of these tools was in compliance with ACM’s Authorship Policy, and all AI-generated text was properly integrated into the paper with appropriate citations.

\bibliographystyle{ACM-Reference-Format}
\balance
\bibliography{mybibfile}

\end{document}